\documentclass[lettersize,journal]{IEEEtran}
\usepackage{amsmath,amsfonts}
\usepackage{algorithmic}
\usepackage{algorithm}
\usepackage{array}
\usepackage[caption=false,font=normalsize,labelfont=sf,textfont=sf]{subfig}
\usepackage{textcomp}
\usepackage{stfloats}
\usepackage{url}
\usepackage{verbatim}
\usepackage{graphicx}
\usepackage{cite}
\usepackage{graphicx}
\usepackage{amsmath}
\usepackage{authblk}
\usepackage[margin=1in]{geometry}
\usepackage{caption}
\usepackage{subcaption}
\usepackage{hyperref}
\usepackage{booktabs}
\usepackage{siunitx}
\usepackage{xcolor}
\usepackage{amssymb}
\usepackage[utf8]{inputenc}
\usepackage[T1]{fontenc}
\usepackage{multirow}  

\begin{document}
	
	\title{MCD-Net: A Lightweight Deep Learning Baseline for Optical-Only Moraine Segmentation}

	\author{Zhehuan Cao, Fiseha Berhanu Tesema,~\IEEEmembership{IEEE member, Ping Fu, Jianfeng Ren, Ahmed Nasr
		}
		
		\thanks{Zhehuan Cao and Jianfeng Ren are with the School of Computer Science, University of Nottingham Ningbo China (UNNC), Ningbo, China}
		\thanks {Ping Fu is with the School of Geographical Sciences, UNNC, Ningbo,China}
		\thanks{Fiseha Berhanu Tesema is with the School of Computer Science and the Nottingham Ningbo China Beacons of Excellence Research and Innovation Institute, UNNC, Ningbo, China.(Corresponding authors: Fiseha Berhanu Tesema e-mail:Fiseha-Berhanu.Tesema@nottinham.edu.cn) and Ping Fu, email:Ping.Fu@nottingham.edu.cn)}
		\thanks{
			Ahmed Nasr is with the Department of Electrical and Electronic Engineering, UNNC, Ningbo, China}
		
	}
	
	\markboth{Journal of \LaTeX\ Class Files,~Vol.~14, No.~8, August~2021}%
	{Shell \MakeLowercase{\textit{et al.}}: A Sample Article Using IEEEtran.cls for IEEE Journals}
	
	
	\maketitle

	\begin{abstract}
		Glacial segmentation is essential for reconstructing past glacier dynamics and evaluating climate-driven landscape change. However, weak optical contrast and the limited availability of high-resolution DEMs hinder automated mapping. This study introduces the first large-scale optical-only moraine segmentation dataset, comprising 3,340 manually annotated high-resolution images from Google Earth covering glaciated regions of Sichuan and Yunnan, China. We develop MCD-Net, a lightweight baseline that integrates a MobileNetV2 encoder, a Convolutional Block Attention Module (CBAM), and a DeepLabV3+ decoder. Benchmarking against deeper backbones (ResNet152, Xception) shows that MCD-Net achieves 62.3\% mean Intersection over Union (mIoU) and 72.8\% Dice coefficient while reducing computational cost by more than 60\%. Although ridge delineation remains constrained by sub-pixel width and spectral ambiguity, the results demonstrate that optical imagery alone can provide reliable moraine-body segmentation. The dataset and code are publicly available at \url{https://github.com/Lyra-alpha/MCD-Net}, establishing a reproducible benchmark for moraine-specific segmentation and offering a deployable baseline for high-altitude glacial monitoring.
	\end{abstract}
	
	\begin{IEEEkeywords} Moraine segmentation, Deep learning, Optical imagery, Lightweight neural networks, CBAM, Glacial geomorphology, Remote sensing. \end{IEEEkeywords}
	
	\section{Introduction}
	
	Glacial landforms are critical archives of Earth's climatic history, with their morphology providing insights into palaeoclimate reconstruction, glacier dynamics, and associated hazards such as glacial lake outburst floods \cite{benn1998,benn2002glaciers,paul2010remote}. Among these landforms, moraine ridges are particularly valuable markers of former glacier extents. Their spatial distribution and morphological characteristics inform reconstructions of past climate variability and ice–climate interactions, while also supporting hazard assessments in alpine regions.
	
	Traditional moraine mapping has relied primarily on field surveys and manual interpretation of aerial photographs or satellite imagery. 
	Although such approaches remain valuable, they are labour-intensive, subjective, and difficult to scale, particularly in remote and high-altitude environments \cite{huggel2004evaluation,chen2016glacial}. 
	To reduce interpretation ambiguity, digital elevation model (DEM)-based analyses have been increasingly adopted to enhance terrain understanding. 
	However, their effectiveness is constrained by the availability, cost, and accuracy of high-resolution DEMs, especially in steep or shadowed alpine terrain where topographic artefacts are common \cite{farr2007shuttle,liu2021evaluating}.
	
	Recent advances in deep learning, particularly convolutional neural networks (CNNs) and encoder–decoder architectures such as U-Net \cite{ronneberger2015unet} and DeepLabV3+ \cite{chen2018deeplab}, have significantly improved automated analysis of remote-sensing imagery. 
	These models have demonstrated strong performance in cryospheric applications including glacier boundary delineation, rock-glacier inventorying, and supraglacial lake detection \cite{zhu2017deep,ma2019deep,zhang2021automatic,hu2023mapping,sun2024tprogi}. 
	By learning hierarchical spatial and contextual features, deep networks offer a scalable alternative to manual geomorphological mapping.
	
	Despite this progress, moraine-specific segmentation remains underdeveloped. 
	Most existing deep-learning studies either treat moraines as secondary classes within broader landform classification tasks or rely heavily on multi-source data fusion involving optical imagery, DEMs, and Synthetic Aperture Radar (SAR) \cite{li2022mornet,rocamora2024multisource}. 
	Such reliance on auxiliary topographic data limits reproducibility and scalability across regions, particularly in high-altitude areas where DEM quality is inconsistent. 
	Moreover, currently available moraine datasets are small, heterogeneous, and lack standardised pixel-level annotations, hindering fair benchmarking and robust generalisation of learning-based methods.
	
	To address these gaps, this study introduces the first large-scale optical-only moraine segmentation dataset, consisting of 3,340 annotated high-resolution images from Sichuan and Yunnan. We propose MCD-Net, a lightweight DeepLabV3+ variant that combines a MobileNetV2 backbone with a Convolutional Block Attention Module (CBAM), offering an efficient and reproducible baseline for optical-only moraine mapping.
	
	This study makes four key contributions:
	\begin{itemize}
		
		\item A new publicly released dataset of 3,340 high-resolution optical-only moraine segmentation images.
		
		\item A compact baseline model, MCD-Net, designed for efficient and reproducible moraine segmentation.
		
		\item A systematic comparison of backbone depth, attention mechanisms, and architectural choices.
		
		\item A robustness evaluation across diverse geomorphological and environmental conditions.
		
	\end{itemize}

	Our objective is to establish a transparent benchmark for moraine segmentation and to assess the feasibility and limitations of optical-only approaches for high-altitude geomorphological analysis.

	\section{Related Work}
	
	\subsection{Classical Geomorphological Mapping}
	Early attempts at moraine identification primarily relied on geomorphological visual interpretation and field surveys. Moraine ridges and depositional bodies were manually delineated through the analysis of aerial photographs, topographic maps, and medium-resolution satellite imagery, including Landsat and SPOT data \cite{huggel2004evaluation,paul2010remote}. These approaches typically focused on diagnostic morphological characteristics such as arcuate ridge geometry, slope asymmetry, surface texture, and valley-parallel spatial alignment.
	
	Despite their value, traditional mapping approaches suffer from several inherent limitations. Manual interpretation is time-consuming, subject to observer bias, and difficult to reproduce consistently across regions and analysts. The subsequent availability of high-resolution optical imagery and three-dimensional visualisation platforms, most notably Google Earth, has partially alleviated these challenges by enabling detailed inspection of fine-scale moraine morphology and terrain context. In particular, the integration of high-resolution Google Earth imagery with interactive 3D viewing has enhanced the identification of subtle moraine features that are difficult to resolve in medium-resolution satellite data. For example, \cite{chen2016glacial} demonstrated that Google Earth–based interpretation facilitated more precise delineation of latero-frontal and hummocky moraines in the southeastern Tibetan Plateau. Nevertheless, such workflows remain heavily dependent on expert judgement, labour-intensive, and unsuitable for large-scale or fully reproducible mapping.

	\subsection{Deep Learning in Glacial Geomorphology}
	Recent advances in deep learning have substantially reshaped the automatic mapping of glacial and periglacial landforms from remote sensing data. Convolutional neural networks (CNNs), particularly encoder--decoder segmentation architectures such as U-Net \cite{ronneberger2015unet} and DeepLabV3+, have become dominant due to their ability to capture multi-scale contextual information and complex spatial patterns \cite{zhu2017deep,ma2019deep}. When combined with data augmentation strategies, these models demonstrate improved robustness and generalisation, especially under limited annotation regimes \cite{shorten2019survey}. Consequently, U-Net-, DeepLab-, and hybrid variants have been widely adopted in cryospheric remote sensing applications \cite{tang2024automatic,periyasamy2022unet,zhang2021automated,lin2023accurate,erharter2022machine,xie2022progressive}.
	
	Within glaciology, deep learning has been successfully applied to tasks such as rock glacier mapping, glacier boundary delineation, and supraglacial feature detection. For example, \cite{hu2023mapping} applied a DeepLabV3+-based framework to rock glacier identification in the West Kunlun Mountains, achieving a training/validation intersection-over-union (IoU) of 0.801 and producing one of the earliest comprehensive regional inventories derived from deep learning methods. Building upon this work, \cite{sun2024tprogi} extended the same methodological framework to a larger-scale automated cataloguing effort across the Tibetan Plateau. Using a DeepLabV3+ model with an Xception71 backbone trained on 4,085 annotated rock glacier samples, their approach achieved encouraging performance (precision 0.55, recall 0.73, F1-score 0.63), although manual post-processing remained necessary for output refinement. Beyond rock glaciers, CNN-based models have also been adopted for glacier outline extraction \cite{baumhoer2019automated} and supraglacial lake detection \cite{zhang2021automatic}, demonstrating the broader applicability of deep learning to cryospheric landform analysis.

	\subsection{Research Gap in Moraine-Specific Identification}
	Despite the rapid growth of deep learning applications in glacial geomorphology, moraine-specific identification remains comparatively underexplored. As highlighted in the recent review of artificial intelligence applications in glacier studies by \cite{nurakynov2024application}, most existing works treat moraines as auxiliary classes within broader landform classification frameworks, rather than as primary targets of investigation. Only a limited number of studies explicitly focus on moraine segmentation.
	
	Among these, the MorNet framework proposed by \cite{li2022mornet} demonstrated the feasibility of deep learning for moraine mapping by integrating multispectral imagery, synthetic aperture radar data, and high-resolution digital elevation models (DEMs). While this multimodal strategy improved performance (F1-score = 52.8, IoU = 35.9) in Himalayan test regions, it also introduced notable limitations. First, reliance on high-resolution DEMs and multi-source inputs constrains scalability and reproducibility, as DEM acquisition is costly and often affected by noise and artefacts in steep or shadowed alpine terrain \cite{farr2007shuttle,liu2021evaluating,rocamora2024multisource}. Second, moraine ridges are typically narrow, fragmented, and visually subtle in optical imagery, frequently obscured by vegetation or exhibiting sub-pixel widths, which complicates annotation and model learning \cite{benn1998}. Furthermore, the small dataset size used in prior studies, such as the 59 annotated polygons in MorNet, limits the robustness and generalisability of the resulting models.
	
	\subsection{Positioning of This Study}
	In response to these gaps, this study aims to establish a reproducible and scalable benchmark for moraine segmentation using optical imagery alone. We propose a new methodological pathway that deliberately avoids dependence on DEMs or multi-source data, focusing instead on high-resolution optical imagery combined with efficient deep learning architectures. By constructing a curated dataset comprising 3,340 annotated moraine images and introducing a lightweight baseline model (MCD-Net), this work directly addresses the limitations of data scarcity, methodological reproducibility, and computational efficiency identified in previous studies. The following section describes the dataset construction process and its key characteristics, which form the foundation of the proposed approach.

	\section{Dataset and Evaluation Metrics}
	
	\subsection{Construction and Coverage}
\begin{figure}[h]
		\centering
		\includegraphics[width=1\linewidth]{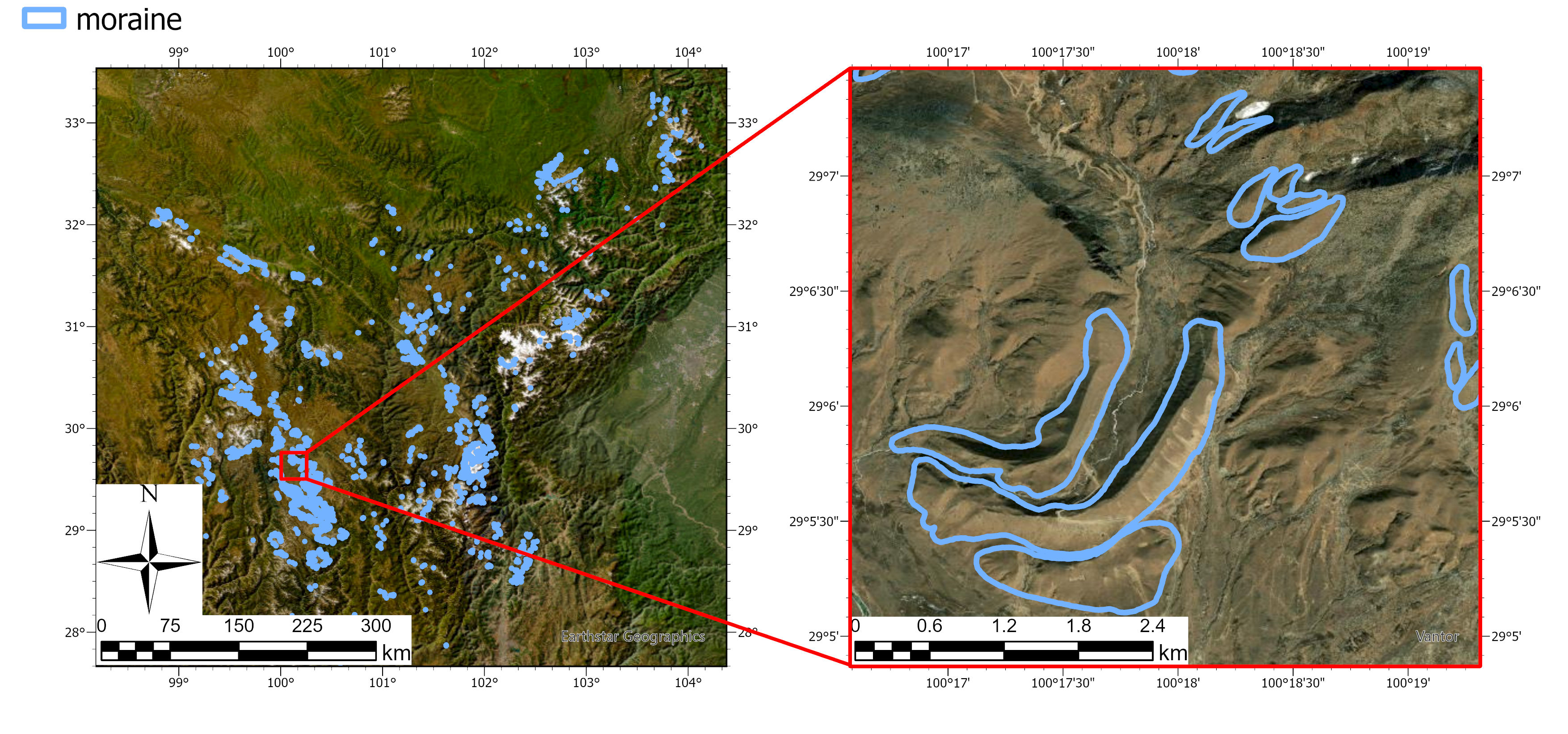}
		\caption{Geographical distribution of moraine areas and representative examples from Sichuan and Yunnan.}
		\label{fig:Moraine_dataset01}
	\end{figure}

	We construct a high-resolution optical dataset dedicated to moraine identification, comprising 3,340 image–mask pairs of $1024 \times 1024$ pixels. 
	Images were sourced from Google Earth Pro, with a spatial resolution of 0.5–2.0~m/pixel and a temporal range from 2020 to 2025. 
	The study area covers glaciated regions of Sichuan and Yunnan, China (26°–32°N, 98°–104°E), including mountain ranges such as Gongga, Que’er, Yulong, and Meili. 
	Elevations span 2,800–5,200~m, encompassing a wide variety of moraine landform types—from cirque and valley moraines to piedmont deposits—under diverse illumination and terrain conditions. 
	Our study area partially overlaps with that of \cite{fu2012glacial}, enabling cross-validation with earlier geomorphological mapping efforts. 
	Figure~\ref{fig:Moraine_dataset01} shows the geographical distribution of the sampled regions and representative moraine environments.
	
	\noindent
	
	Unlike the MorNet collection \cite{li2022mornet}, which contains only 59 annotated polygons and depends on multi-source data fusion (optical, Digital Elevation Model—DEM, and Synthetic Aperture Radar—SAR), our MCD Dataset comprises 3,340 high-resolution, optical-only image–mask pairs. 
	This represents a 56-fold increase in data volume and, importantly, removes dependence on costly high-precision DEMs. 
	By focusing exclusively on optical imagery, the MCD dataset enhances reproducibility, scalability, and accessibility for the broader research community working on automated moraine segmentation.
	
	\subsection{Annotation Protocol}
	Three trained geomorphologists independently annotated moraine bodies and ridges, with majority voting used to reconcile disagreements. Initially, annotations distinguished background (0), moraine body (1), and moraine ridge (2). However, ridges represented only 0.2\% of pixels and showed significant annotation ambiguity (average $\pm 2$ pixels across annotators). To improve consistency, ridge pixels were merged into the moraine body class, producing a binary segmentation task (background vs.\ moraine body).
	
\begin{figure}[h]
		\centering
		\includegraphics[width=1\linewidth]{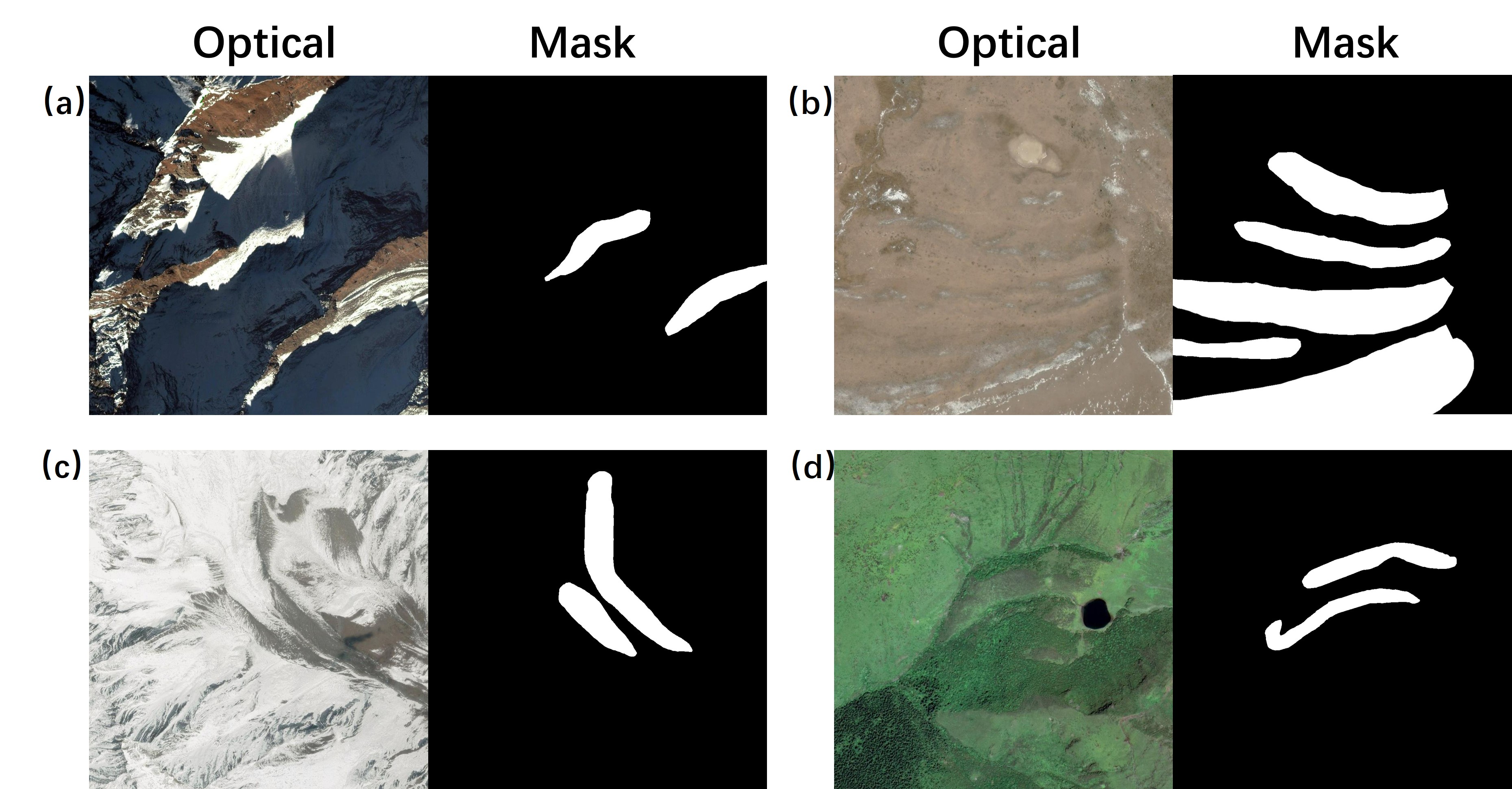}
		\caption{Representative dataset examples under varying conditions: (a) strong shadows, (b) low-contrast rocky surfaces, (c) glacier-covered moraine deposits, (d) vegetation-covered moraines.}
		\label{fig:dataset_example02}
	\end{figure}

	We also evaluated the consistency between our annotations and an earlier version of moraine mapping from \cite{fu2012glacial} in the overlapping regions. The Intersection over Union (IoU) between the two versions was 0.313, reflecting both the improved resolution of modern imagery and the evolution of annotation criteria over the past decade. This relatively low agreement quantitatively demonstrates the high annotation uncertainty of ridge structures, further supporting our decision to merge them into the moraine body category.
	Figure~\ref{fig:dataset_example02} presents sample examples and annotations from the dataset, illustrating the challenging conditions such as shadows, low-contrast surfaces, and vegetation cover that our model must overcome.

	\subsection{Class Imbalance and Scale Diversity}
\begin{figure}[h]
		\centering
		\includegraphics[width=1\linewidth]{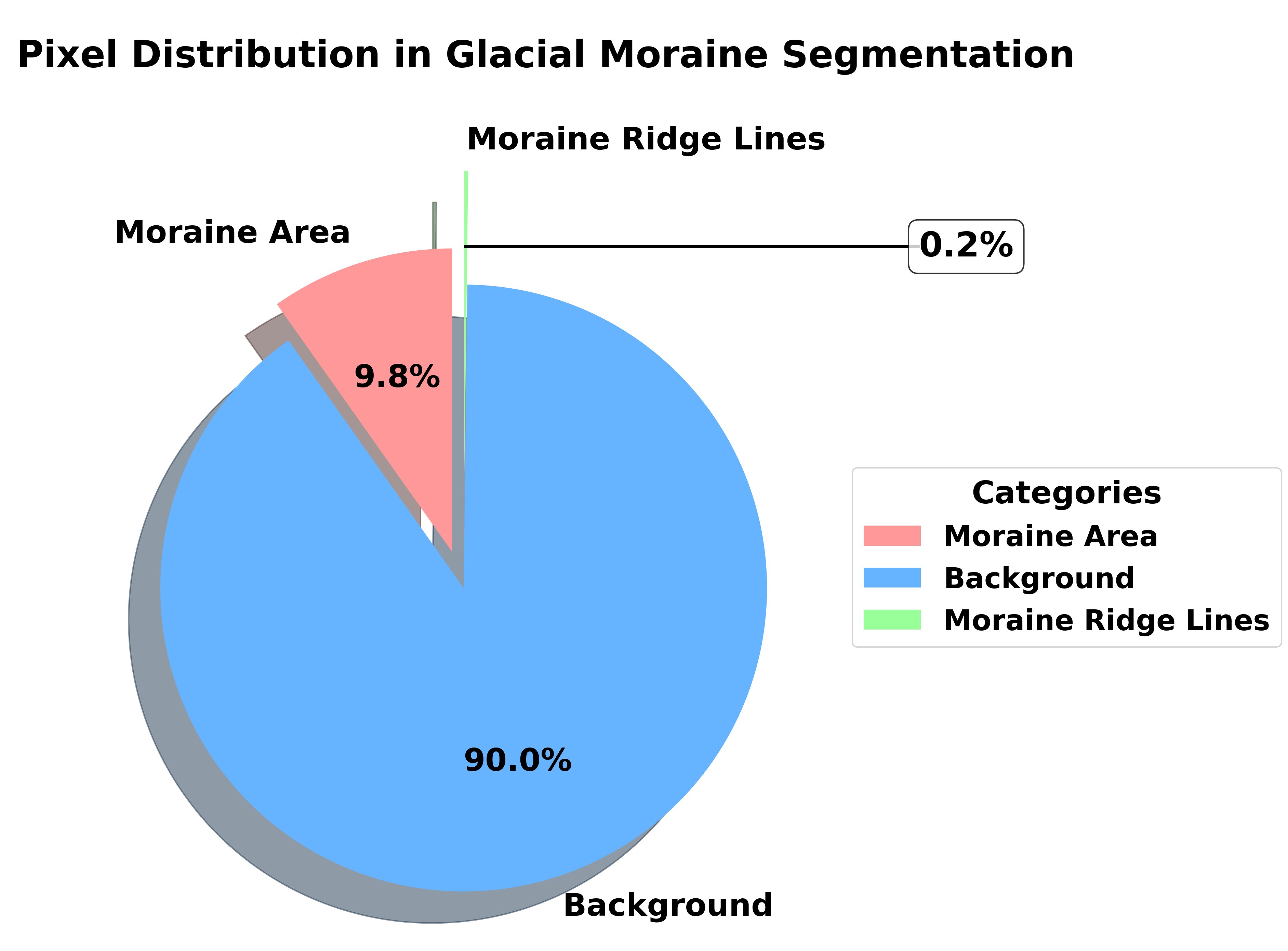}
		\caption{Pixel distribution across classes.}
		\label{fig:pixel_distribution}
	\end{figure}

	Pixel-level statistics confirm extreme class imbalance: background pixels account for $\sim 90.0\%$, moraine body for 9.8\%, and ridges (excluded) for only 0.2\% (Fig.~\ref{fig:pixel_distribution}). Moraines also show a long-tailed size distribution: some cover only a few hundred pixels, while others exceed hundreds of thousands of pixels. This variability necessitates models capable of handling multi-scale targets.
	
\begin{figure}[h]
		\centering
		\includegraphics[width=1\linewidth]{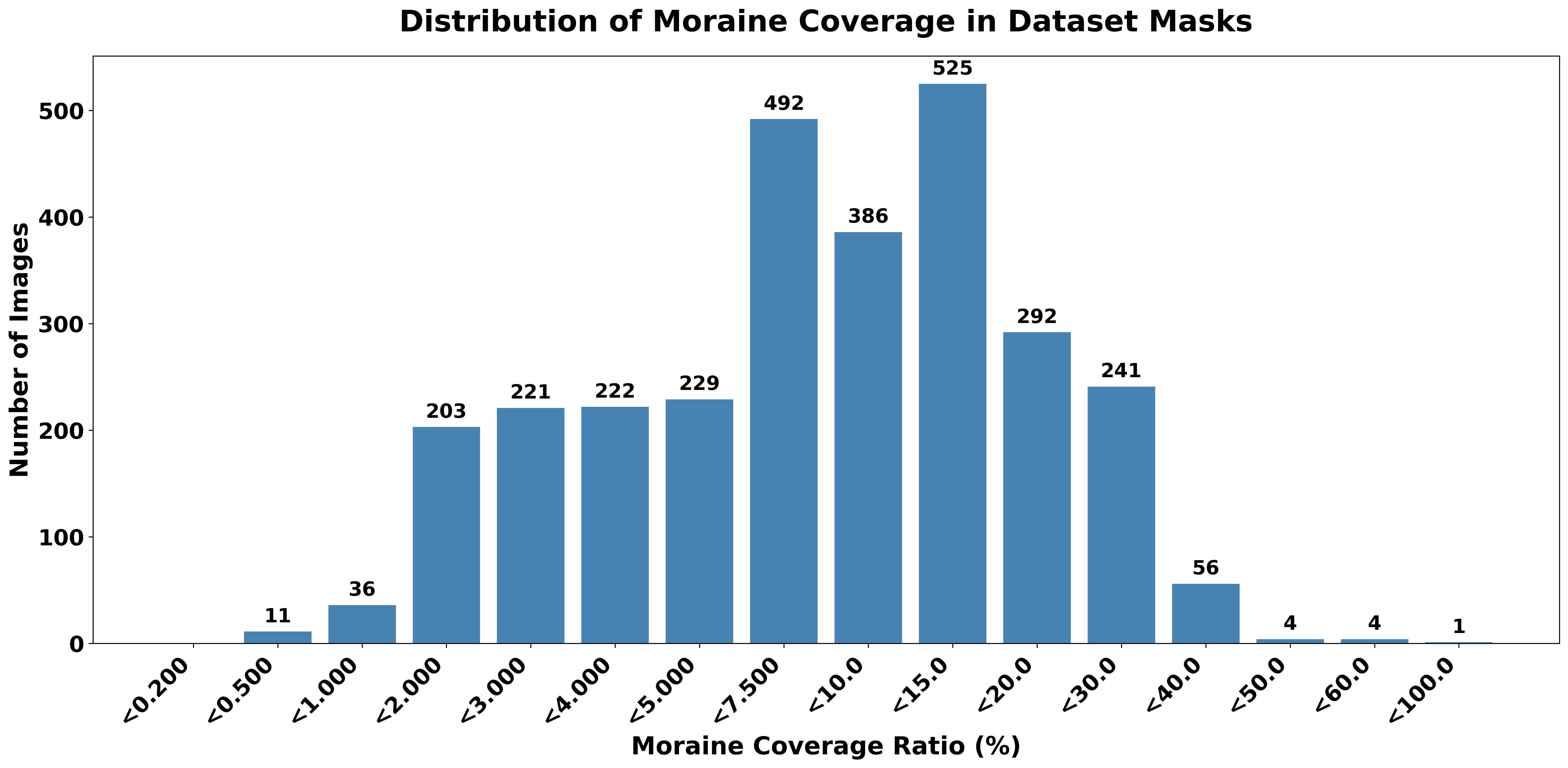}
		\caption{Histogram of moraine sizes (in pixels).}
		\label{fig:moraine_area_histogram}
	\end{figure}

	The moraine size distribution histogram (Fig.~\ref{fig:moraine_area_histogram}) further reveals the diversity characteristics of the dataset. The figure shows that moraine coverage in the dataset ranges from 0.2\% to over 60\%, exhibiting a clear long-tail distribution. This indicates that the dataset contains both small arcuate ridges occupying only a few hundred pixels and large valley moraines covering hundreds of thousands of pixels. Such extreme scale differences require models to possess the capability to handle multi-scale targets.
	
	Given that ridge morphology often appears as sub-pixel width (average less than 3 pixels) in optical imagery and exhibits significant annotation ambiguity (approximately ±2 pixels inter-annotator variation) \cite{ma2019deep}, this study optimized the annotation scheme to a binary classification problem by merging the "moraine ridge" category into the "moraine body" category. This decision not only ensures annotation consistency but, more importantly, aligns the task objectives with the practical discriminative capabilities of current-resolution optical imagery \cite{paul2010remote}.

	\subsection{Partitioning and Preprocessing}
	The dataset is randomly split into training (2{,}630 images) and test (293 images) subsets in a 9:1 ratio. 
	Coverage across multiple valleys ensures geographical diversity in both partitions. 
	Preprocessing includes normalising pixel values to $[0,1]$ and converting arrays to channel-first tensors for PyTorch implementation.
	Table~\ref{tab:dataset_stats} summarises the dataset composition and pixel-level proportions for each category.
	
	\begin{table*}[t]
		\centering
		\caption{Dataset composition and statistics. Percentages are computed over all annotated pixels.}
		\label{tab:dataset_stats}
		\begin{tabular}{lccc}
			\toprule
			Category & Number of Images & Pixel Proportion (\%) & Notes \\
			\midrule
			Background & 3340 & 90.0 & Non-moraine terrain \\
			Moraine Body & 2923 & 9.8 & Includes merged ridges \\
			Ridge (excluded) & 417 & 0.2 & Sub-pixel features \\
			\midrule
			Training Set & 2630 & --- & 9 parts \\
			Test Set & 293 & --- & 1 part \\
			\bottomrule
		\end{tabular}
	\end{table*}
	
	As shown in Table~\ref{tab:dataset_stats}, background pixels dominate the dataset, with moraine bodies accounting for only 9.8\% of the total area.
	This imbalance underscores the need for class-weighted losses and evaluation metrics that remain robust under extreme pixel disparities.
	
	\subsection{Evaluation Protocol and Metrics}
	To ensure reproducibility and facilitate fair benchmarking on the proposed MCD dataset, we establish a standard evaluation protocol and adopt widely used metrics for semantic segmentation. 
	All metrics are computed at the pixel level, treating the task as binary classification between moraine body and background.
	
	For each image, the predicted segmentation map is compared against the manually annotated ground truth using Intersection over Union (IoU), mean IoU (mIoU), Dice coefficient, Precision (Prec), Recall (Rec), and Pixel Accuracy (PA) as follows:
	\begingroup
	\setlength{\arraycolsep}{2pt} 
	\begin{eqnarray}
		\mathrm{IoU}  &=& \frac{TP}{TP+FP+FN}, \\
		\mathrm{mIoU} &=& \frac{1}{C}\sum_{c=1}^{C}\frac{TP_c}{TP_c+FP_c+FN_c}, \\
		\mathrm{Dice} &=& \frac{2TP}{2TP+FP+FN}, \\
		\mathrm{Prec} &=& \frac{TP}{TP+FP}, \\
		\mathrm{Rec}  &=& \frac{TP}{TP+FN}, \\
		\mathrm{PA}   &=& \frac{TP+TN}{TP+TN+FP+FN}.
	\end{eqnarray}
	\endgroup
	
	where $TP$, $FP$, $TN$, and $FN$ denote the number of true positives, false positives, true negatives, and false negatives, respectively, and $C$ represents the number of classes ($C=2$ in this dataset). 
	These metrics collectively quantify model performance in terms of overlap quality, boundary accuracy, and pixel-level reliability.
	
	We recommend mean IoU and Dice coefficient as the primary indicators for future benchmarking on this dataset, as they provide balanced evaluation of segmentation quality under class imbalance. 
	The inclusion of multiple auxiliary metrics (Precision, Recall, and Pixel Accuracy) supports more detailed comparison among models and configurations.

	\section{Methodology: The MCD-Net Architecture}
	
	This section introduces the proposed MCD-Net, a lightweight architecture tailored for moraine segmentation from optical imagery. The design integrates a MobileNetV2 backbone for efficiency, a Convolutional Block Attention Module (CBAM) for feature refinement, and the DeepLabV3+ head for multi-scale segmentation.
	
	\subsection{Motivation for a Lightweight Baseline}
	State-of-the-art backbones such as ResNet152 and Xception achieve strong feature extraction but come with high computational cost and risk of overfitting when training on limited domain-specific datasets such as ours. By contrast, MobileNetV2 employs depthwise separable convolutions and inverted residual blocks, offering a compact parameterisation with high efficiency. These properties make it well suited for scalable or resource-constrained deployments (e.g., UAV-based field monitoring), while still delivering competitive segmentation accuracy.
	
	\begin{figure}[!t]
		\centering
		\includegraphics[width=1\linewidth]{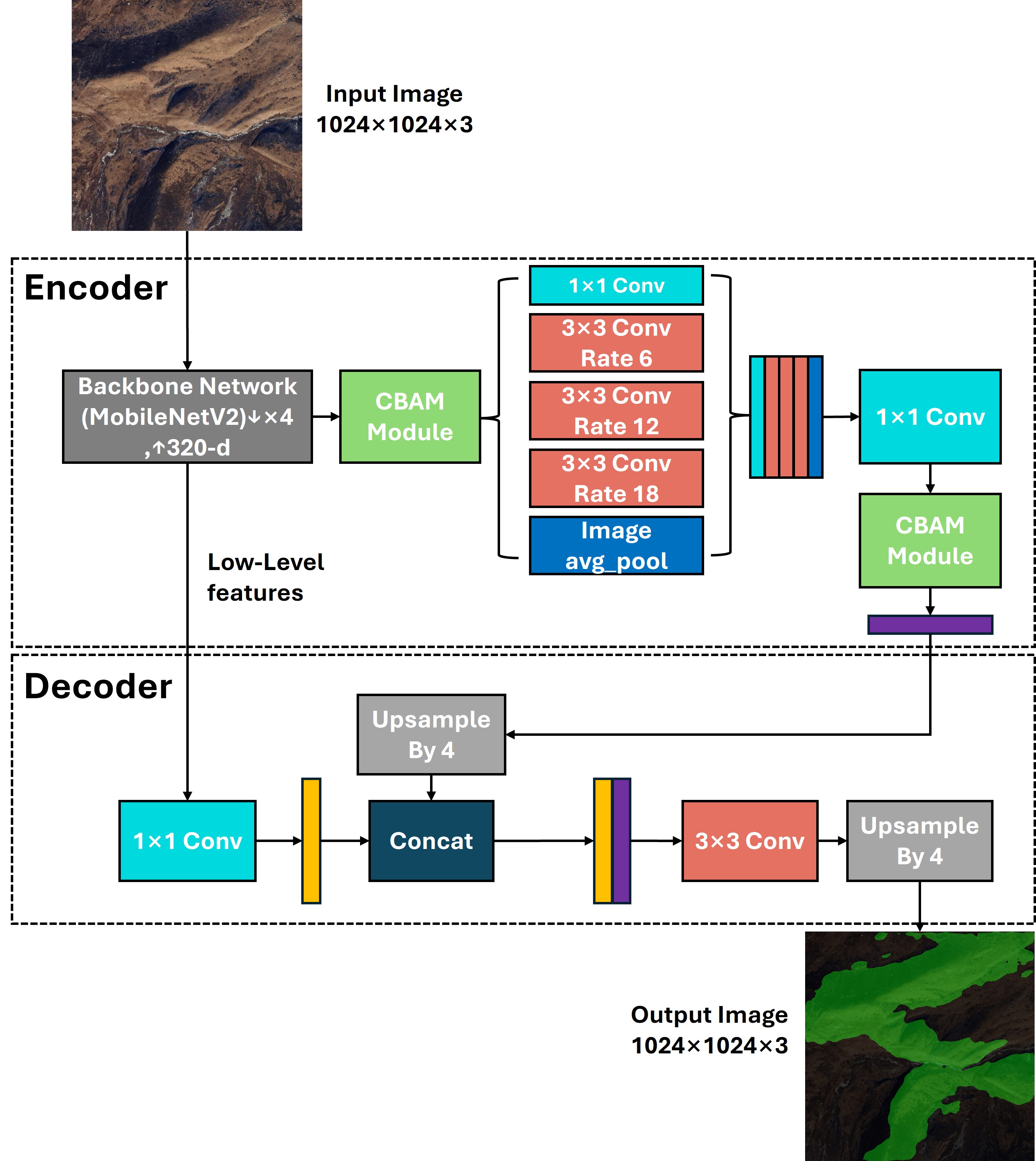}
		\caption{Proposed \textbf{MCD-Net} architecture. The pipeline consists of a MobileNetV2 backbone, a CBAM block for feature refinement, ASPP for multi-scale context, and a DeepLabV3+ decoder.}
		\label{fig:arch_diagram}
	\end{figure}
	
	\subsection{Architecture Overview}
	The overall architecture follows an encoder-decoder paradigm. Input images $I \in \mathbb{R}^{H \times W \times 3}$ are first processed by the MobileNetV2 backbone to produce feature maps:
	\begin{equation}
		F_{\text{base}} = \mathcal{B}(I),
	\end{equation}
	where $\mathcal{B}$ denotes the backbone transformation.
	
	The overall architecture of our proposed MCD-Net is illustrated in Figure~\ref{fig:arch_diagram}, which shows the integration of MobileNetV2 backbone, CBAM attention module, and DeepLabV3+ decoder components.
	
	To enhance feature discrimination under subtle spectral contrasts, CBAM is applied to refine $F_{\text{base}}$, yielding attention-enhanced features $F_{\text{att}}$:
	\begin{equation}
		F_{\text{att}} = \mathcal{M}_s \big( \mathcal{M}_c(F_{\text{base}}) \big) \odot F_{\text{base}},
	\end{equation}
	where $\mathcal{M}_c$ and $\mathcal{M}_s$ denote channel and spatial attention maps, and $\odot$ is element-wise multiplication.
	
	The refined features are then passed through the Atrous Spatial Pyramid Pooling (ASPP) module for multi-scale context aggregation:
	\begin{equation}
		F_{\text{aspp}} = \mathcal{A}(F_{\text{att}}),
	\end{equation}
	where $\mathcal{A}$ represents the ASPP transformation.
	
	Finally, a decoder $\mathcal{D}$ upsamples the contextual features to produce the binary segmentation map:
	\begin{equation}
		\hat{Y} = \mathcal{D}(F_{\text{aspp}}), \quad \hat{Y} \in \{0,1\}^{H \times W}.
	\end{equation}

	\subsection{Convolutional Block Attention Module (CBAM)}
\begin{figure}[h]
		\centering
		\includegraphics[width=1\linewidth]{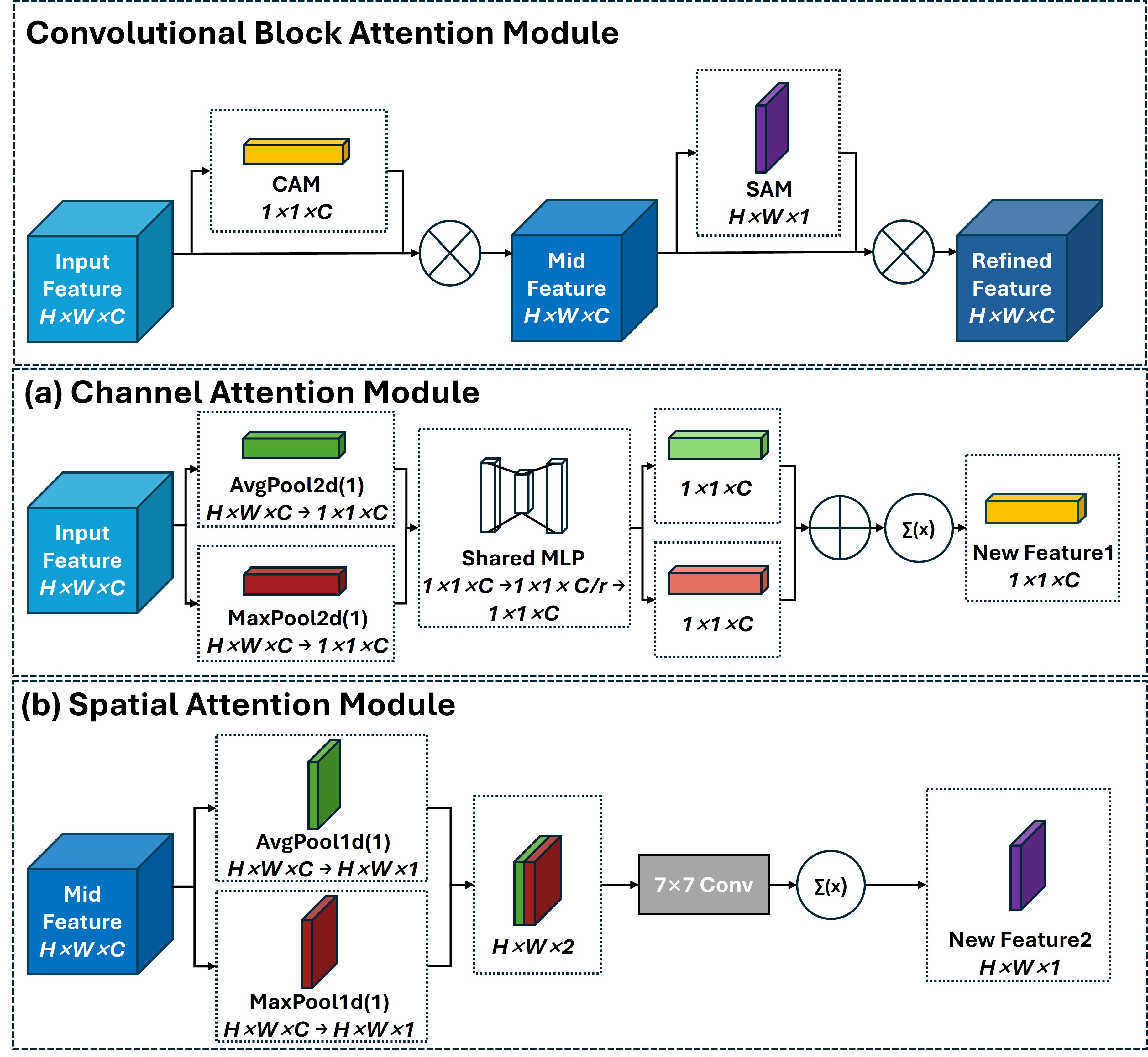}
		\caption{The CBAM module, consisting of (a) channel attention using pooled descriptors and MLP layers, and (b) spatial attention using concatenated pooling maps and convolution.}
		\label{fig:cbam_module}
	\end{figure}

	The detailed structure of the Convolutional Block Attention Module (CBAM) is illustrated in Figure~\ref{fig:cbam_module}, showing both the channel attention and spatial attention components that work in sequence to refine feature representations.
	CBAM sequentially applies channel and spatial attention to highlight informative features:
	\begin{align}
		\mathbf{M}_c(F) &= \sigma \big( \text{MLP}(\text{AvgPool}(F)) + \text{MLP}(\text{MaxPool}(F)) \big), \\
		\mathbf{M}_s(F) &= \sigma \big( f^{7 \times 7}([\text{AvgPool}(F); \text{MaxPool}(F)]) \big),
	\end{align}
	where $\sigma$ is the sigmoid function, MLP denotes a two-layer perceptron with shared weights, $f^{7 \times 7}$ is a convolution with kernel size $7 \times 7$, and $[\cdot;\cdot]$ denotes concatenation. 
	
	The final refined feature map is:
	\begin{equation}
		F_{\text{att}} = \mathbf{M}_s(F) \odot \mathbf{M}_c(F) \odot F.
	\end{equation}
	
	\subsection{Training Settings}
	All models are trained using the AdamW optimiser with an initial learning rate $1 \times 10^{-4}$, weight decay of $1 \times 10^{-4}$, and a cosine annealing learning-rate schedule. 
	We employ a cross-entropy loss
	\begin{equation}
		\mathcal{L}_{\text{CE}} = - \sum_{c \in \{0,1\}} w_c \, y_c \log(p_c),
	\end{equation}
	where $y_c$ and $p_c$ denote the ground-truth label and predicted probability for class $c$. 
	Equal class weights ($w_{\text{bg}} = w_{\text{moraine}} = 0.5$) are adopted to maintain stable optimisation and to avoid over-penalising the minority class under noisy annotations. 
	Although the dataset exhibits strong pixel-level class imbalance, we rely on overlap-based evaluation metrics (mIoU and Dice), which are robust to skewed class distributions, and on architectural feature refinement through ASPP and attention mechanisms rather than aggressive loss reweighting.
	
	Data augmentation includes random scaling (0.5–2.0), horizontal and vertical flipping, rotation ($\pm30^{\circ}$), and Gaussian blur to improve generalisation. 
	Training is conducted for up to 200 epochs on an NVIDIA RTX~5060~Ti GPU, with early stopping applied using a patience of 15 epochs.

	\section{Experiments}
	\label{sec:experiments}
	
	This section presents the experimental design, baseline configurations, and evaluation of the proposed MCD-Net on the newly developed MCD dataset. All experiments were conducted to ensure reproducibility and to provide a transparent foundation for benchmarking future moraine segmentation studies.
	
	\subsection{Experimental Setup}
	All experiments are implemented in \textit{PyTorch}, using input tiles of $1024\times1024$ pixels. 
	The AdamW optimiser is adopted with an initial learning rate of $1\times10^{-4}$, weight decay of $1\times10^{-4}$, and cosine-annealing learning schedule. 
	Training is performed with a batch size of 16 and early stopping (patience = 15 epochs). 
	
	To enhance model generalisation, data augmentation includes random scaling (0.5–2.0), horizontal and vertical flipping, rotation ($\pm30^{\circ}$), and Gaussian blur. 
	All experiments are conducted on an NVIDIA RTX~5060~Ti GPU (16~GB memory). 
	Each model is trained for up to 200 epochs, with the best checkpoint selected based on mean Intersection over Union (mIoU) on the validation split. Model computational complexity (FLOPs) and parameter counts are evaluated using the \texttt{ptflops} toolkit with $1024\times1024$ input resolution to ensure reproducible benchmarking.

	\subsection{Ablation Studies}
	To evaluate the contribution of individual components, we conducted ablation experiments focusing on (1) backbone selection and (2) the effect of integrating the Convolutional Block Attention Module (CBAM)~\cite{woo2018cbam}. 
	The quantitative results are summarised in Table~\ref{tab:backbone_cbam}.
	
	To better understand how architectural design choices influence performance, detailed ablation studies are presented next.
	
	\begin{table*}[t]
		\centering
		\caption{Ablation study results: performance comparison of different backbones with and without CBAM on the test set. 
			Model complexity is reported in terms of parameter count (M) and computational cost (GFLOPs) for $1024 \times 1024$ input resolution.}
		\label{tab:backbone_cbam}
		\begin{tabular}{lccccccc}
			\toprule
			\multirow{2}{*}{Model} & \multicolumn{2}{c}{Complexity} & \multicolumn{5}{c}{Performance Metrics} \\
			\cmidrule(lr){2-3} \cmidrule(lr){4-8}
			& Params (M) & GFLOPs & mIoU (\%) & Recall (\%) & Precision (\%) & Dice (\%) & Pixel Acc. (\%) \\
			\midrule
			MobileNetV2 & 5.81 & 105.7 & 61.9 & 69.7 & 76.3 & 72.3 & 91.4 \\
			MobileNetV2 + CBAM (\textbf{MCD-Net}) & 5.83 & 105.7 & \textbf{62.3} & \textbf{70.9} & \textbf{75.2} & \textbf{72.8} & \textbf{91.2} \\
			\midrule
			ResNet152 & 74.99 & 433.9 & 60.2 & 70.0 & 71.7 & 70.8 & 90.1 \\
			ResNet152 + CBAM & 76.03 & 433.9 & 59.8 & 71.5 & 69.9 & 70.7 & 89.2 \\
			\midrule
			Xception & 54.71 & 333.7 & 56.5 & 63.5 & 71.3 & 66.2 & 90.2 \\
			Xception + CBAM & 55.30 & 333.7 & 56.7 & 63.9 & 71.0 & 66.4 & 90.1 \\
			\bottomrule
		\end{tabular}
	\end{table*}
	
	\paragraph{Backbone analysis.}
	Table~\ref{tab:backbone_cbam} summarises both the computational complexity and segmentation performance of all evaluated architectures. 
	The proposed MCD-Net (MobileNetV2 + CBAM) achieves the best overall results, with an mIoU of 62.3\% and a Dice coefficient of 72.8\%, while maintaining exceptional efficiency at only 5.83~million parameters and 105.7~GFLOPs. 
	By comparison, deeper backbones such as ResNet152 (60.2\% mIoU, 70.8\% Dice) and Xception (56.5\% mIoU, 66.2\% Dice) demand substantially higher computational resources—up to 74.99~million parameters and 433.9~GFLOPs for ResNet152—yet yield lower accuracy due to overfitting on the limited and visually complex moraine dataset.
	These results confirm that lightweight backbones, when enhanced with targeted attention refinement, can extract subtle geomorphological features more effectively than heavier architectures. 
	MobileNetV2’s depthwise separable convolutions capture fine-scale textures and boundary variations of moraine bodies while preserving generalisation stability, establishing MCD-Net as a reproducible and efficient optical-only baseline for moraine segmentation research.
	
	\paragraph{Effect of attention.}
	The integration of the Convolutional Block Attention Module (CBAM) yields a measurable yet architecture-dependent improvement. 
	For the lightweight MobileNetV2 backbone, adding CBAM enhances segmentation performance by +0.35\% in mIoU and +0.47\% in Dice coefficient, resulting in the proposed MCD-Net. 
	This demonstrates that compact models benefit from selective channel–spatial attention, which helps emphasise moraine-relevant spectral and morphological cues while suppressing noisy background textures. 
	In contrast, deeper architectures such as ResNet152 and Xception exhibit negligible or slightly negative performance shifts when CBAM is applied, indicating redundancy in their already high representational capacity. 
	These findings confirm that attention mechanisms are most beneficial for lightweight backbones with limited feature expressiveness, serving as a targeted enhancement rather than a universal solution for deep architectures.
	
	\subsection{Cross-Region Generalization Test}
	\label{subsec:cross_region}
	
\begin{figure}[t]
		\centering
		\includegraphics[width=1\linewidth]{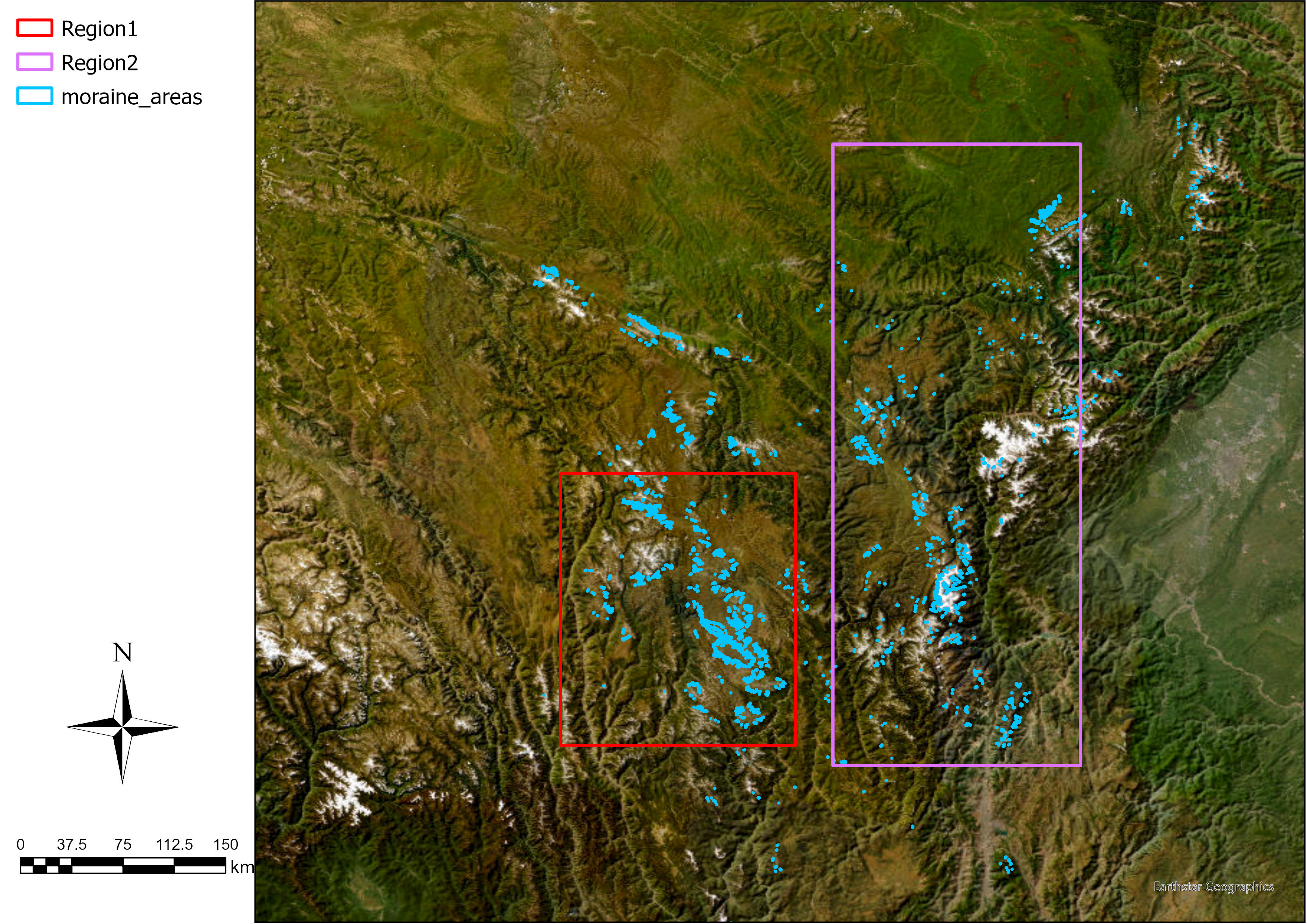}
		\caption{Geographical partition for cross-region evaluation. Region~1 (red) and Region~2 (purple) are non-overlapping rectangular areas within Sichuan; moraine distribution is shown in blue. Region~1 spans 98.937$^{\circ}$E--100.730$^{\circ}$E and 28.491$^{\circ}$N--30.565$^{\circ}$N; Region~2 spans 101.015$^{\circ}$E--102.907$^{\circ}$E and 28.336$^{\circ}$N--33.079$^{\circ}$N.}
		\label{fig:region_partition}
	\end{figure}

	To quantify spatial domain shift within the Sichuan subset, we conduct a bidirectional cross-region evaluation using two non-overlapping rectangular regions: Region~1 (approximately 1{,}237 images) and Region~2 (approximately 1{,}103 images). The geographic partition is illustrated in Fig.~\ref{fig:region_partition}. This controlled split produces two comparably sized subsets and allows us to evaluate how performance changes when the test distribution differs from the training distribution.
	
	We follow the same training configuration as in Sec.~\ref{sec:experiments} (optimizer, learning rate schedule, augmentation, and stopping criterion) and evaluate two settings: (A) train on Region~1 and test on Region~2, and (B) train on Region~2 and test on Region~1. For reference, we also report within-region performance (training and testing on the same region using the same protocol).
	
	\begin{table*}[t]
		\centering
		\caption{Bidirectional cross-region evaluation within Sichuan. $\Delta$mIoU denotes the drop relative to within-region testing.}
		\label{tab:cross_region_results}
		\begin{tabular}{lcccccc}
			\toprule
			\textbf{Experiment} & \textbf{mIoU (\%)} & \textbf{mRecall (\%)} & \textbf{mPrecision (\%)} & \textbf{mF1 (\%)} & \textbf{Acc. (\%)} & \textbf{$\Delta$mIoU} \\
			\midrule
			Train on Region~1 (within-region) & 60.28 & 66.64 & 76.95 & 70.22 & 92.02 & -- \\
			\textbf{Region~1 $\rightarrow$ Region~2} & 53.31 & 59.36 & 68.42 & 61.81 & 89.78 & \textbf{-6.97} \\
			\midrule
			Train on Region~2 (within-region) & 62.69 & 71.51 & 75.45 & 73.25 & 91.32 & -- \\
			\textbf{Region~2 $\rightarrow$ Region~1} & 52.01 & 57.45 & 66.65 & 59.62 & 90.25 & \textbf{-10.68} \\
			\bottomrule
		\end{tabular}
	\end{table*}
	
	As shown in Table~\ref{tab:cross_region_results}, cross-region testing leads to a consistent decrease of 7.0--10.7 points in mIoU compared with within-region evaluation, indicating a non-trivial spatial domain shift between the two areas. The drop is accompanied by reductions in recall and F1, suggesting that part of the shift manifests as increased false negatives when moraines appear under different illumination, background composition, or mosaicking characteristics. Nevertheless, the model maintains mIoU above 52\% in both directions, supporting its use as a baseline while highlighting that improved cross-region robustness remains an open challenge for optical-only moraine mapping.

	\subsection{Qualitative and Attention Results}
\begin{figure}[t]
		\centering
		\includegraphics[width=0.95\linewidth]{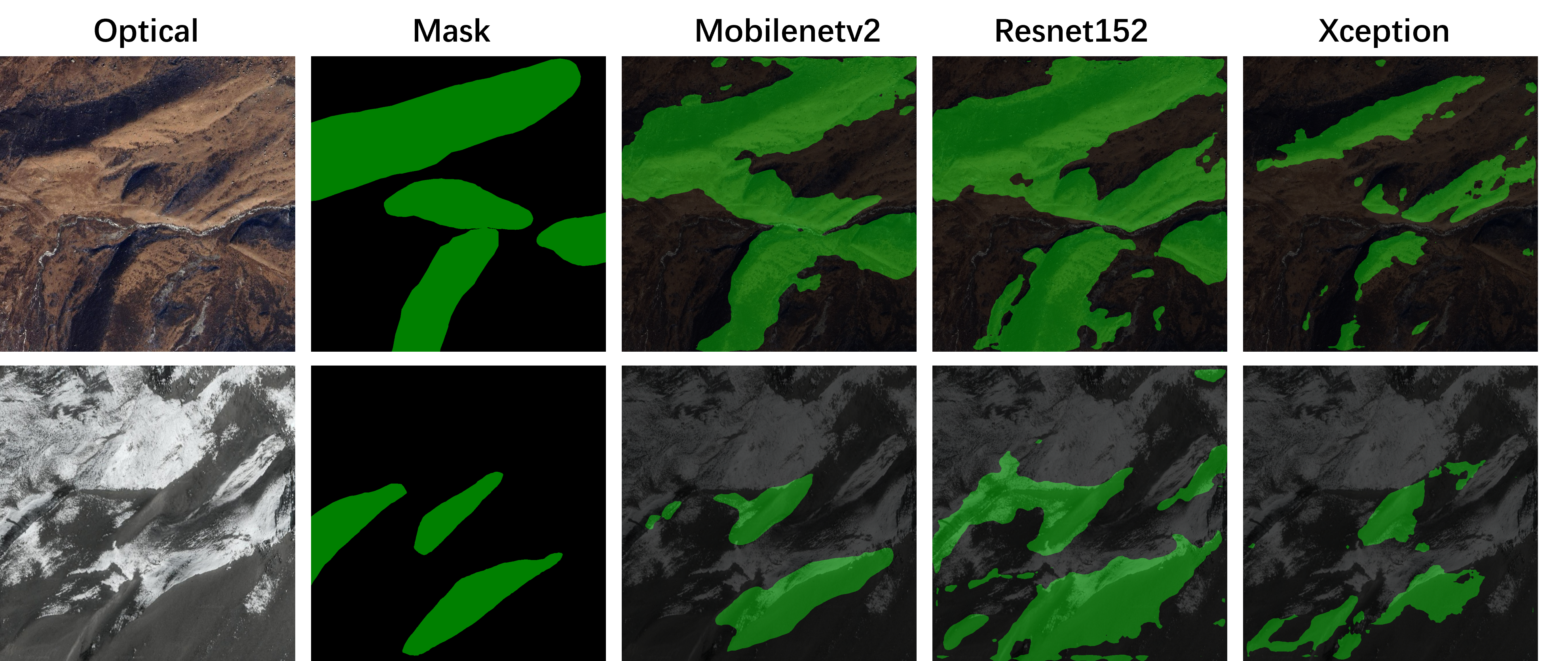}
		\caption{Representative segmentation results. MCD-Net (MobileNetV2) improves ridge continuity and suppresses false positives in shadowed terrain.}
		\label{fig:results}
	\end{figure}

	Figure~\ref{fig:results} illustrates representative segmentation results across different terrain types. 
	MCD-Net using MobileNetV2 produces smoother and more complete moraine boundaries compared to baseline architectures, particularly under shadowed or low-contrast illumination. 
	ResNet152 and Xception often confuse moraine deposits with adjacent debris or vegetation, resulting in false positives.
	
\begin{figure*}[t]
		\centering
		\includegraphics[width=0.9\linewidth]{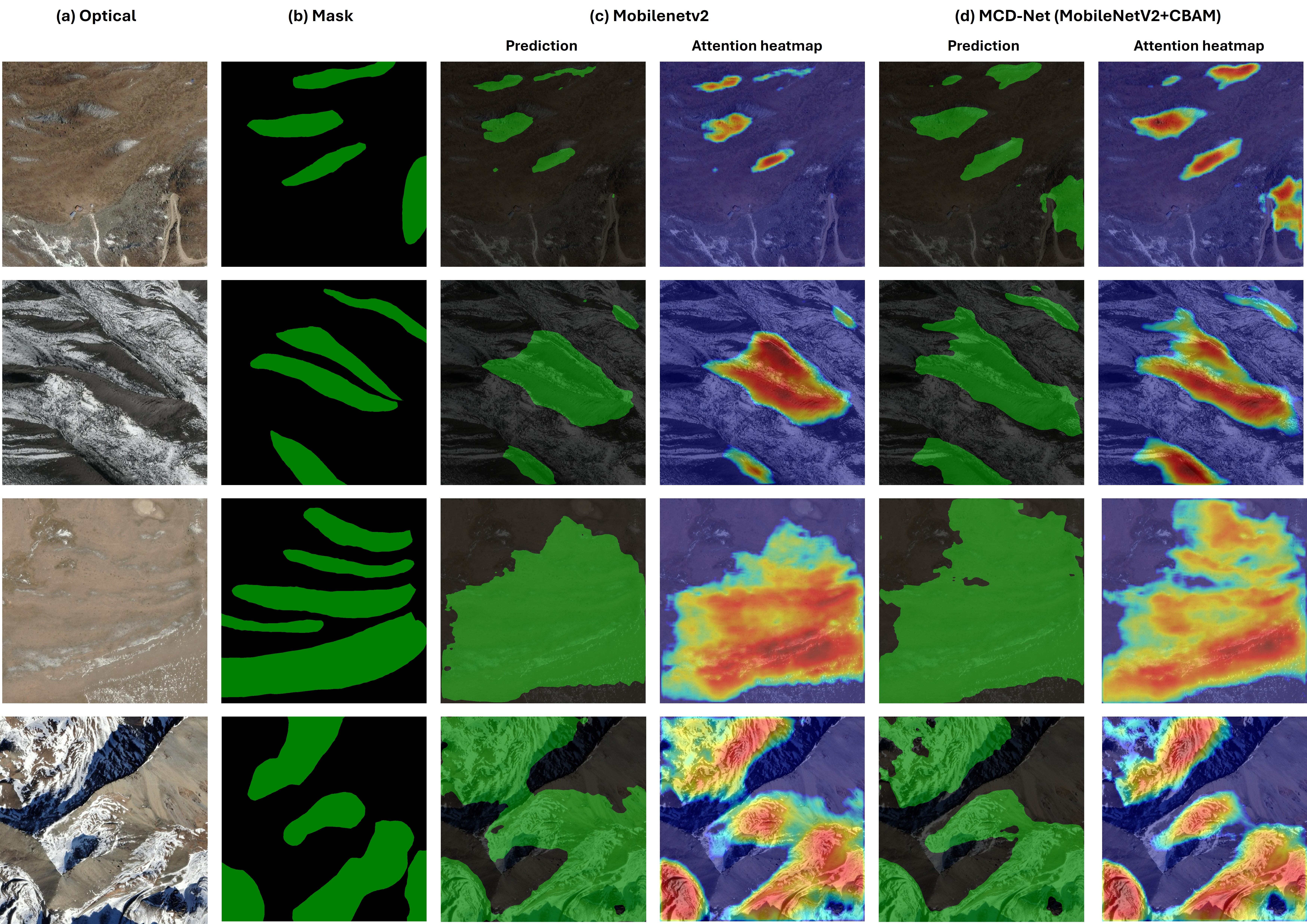}
		\caption{Grad-CAM visualisation of CBAM attention. 
			(a) Original image; (b) Ground truth mask; (c) MobileNetV2 activations; (d) MCD-Net activations. 
			CBAM enhances focus on moraine structures and reduces noise from surrounding terrain.}
		\label{fig:attention_maps}
	\end{figure*}

	To interpret how attention enhances feature discrimination, Figure~\ref{fig:attention_maps} visualises Grad-CAM activations for both MobileNetV2 and MCD-Net. 
	The CBAM-augmented model exhibits focused activation on moraine regions while suppressing irrelevant background textures, demonstrating the interpretability and efficiency of attention in guiding lightweight segmentation networks.
	
	The experimental results validate the robustness and efficiency of MCD-Net for optical-only moraine segmentation. 
	Its lightweight architecture achieves competitive accuracy compared to deep backbones while requiring fewer parameters and computations. 
	These findings establish MCD-Net as a practical and reproducible baseline for future research on moraine recognition and geomorphological segmentation. 
	The following section compares MCD-Net against representative state-of-the-art methods to further contextualise its performance in the broader literature.

	\section{Comparison with Existing Works}
	\label{sec:comparison}
	
	Although deep learning has been widely applied in glacier mapping or rock glacier, dedicated moraine segmentation remains rare. 
	Table~\ref{tab:related_works} summarises representative prior efforts and highlights how this study differs in scope, dataset design, and methodological framework.
	
	\textbf{(1) Task focus.}  
	Most previous works targeted glacier fronts \cite{baumhoer2019automated}, rock glaciers \cite{sun2024tprogi}, or supraglacial lakes \cite{zhang2021automatic}. 
	These studies treated moraines as auxiliary classes within broader landform mapping. 
	Our study isolates moraine segmentation as an independent task, enabling the model to capture its unique depositional morphology and spectral contrast.
	
	\textbf{(2) Dataset scale and diversity.}  
	The MorNet framework \cite{li2022mornet} used only 59 annotated polygons from multi-source data, while our optical-only MCD dataset contains 3,340 annotated image–mask pairs covering diverse moraine types across Sichuan and Yunnan. 
	This expansion improves representativeness and benchmarking reliability.
	
	\textbf{(3) Input modality and reproducibility.}  
	Most existing approaches depend on digital elevation or radar data \cite{rocamora2024multisource, erharter2022machine, xie2022glaciernet2}, which restrict applicability in remote terrain. 
	Our work demonstrates that purely optical imagery can achieve competitive segmentation performance without auxiliary modalities.
	
	\textbf{(4) Architectural distinction.}  
	While prior deep networks typically rely on heavy backbones (e.g., ResNet, Xception), we propose a lightweight MobileNetV2 backbone integrated with a Convolutional Block Attention Module (CBAM). 
	This design achieves state-of-the-art accuracy with a small computational footprint suitable for UAV-based monitoring.
	
	\begin{table*}[t]
		\centering
		\caption{Comparison between the proposed study and representative deep-learning works in cryospheric geomorphology.}
		\label{tab:related_works}
		\begin{tabular}{p{2.4cm}p{2 cm}p{3 cm}p{5cm}}
			\toprule
			\textbf{Study} & \textbf{Target} & \textbf{Data Source} & \textbf{Remarks} \\
			\midrule
			Baumhoer et al. \cite{baumhoer2019automated} & Glacier fronts & Optical (Sentinel-2) & Multi-class segmentation \\
			Sun et al. \cite{sun2024tprogi} & Rock glaciers & Optical + DEM & Plateau-scale inventory (TProGI) \\
			Li et al. \cite{li2022mornet} & Moraines & Optical + DEM + SAR & MorNet; 59 annotated polygons \\
			Erharter et al. \cite{erharter2022machine} & Rock glaciers & DEM & Topography-based ML classification \\
			\textbf{This study} & \textbf{Moraines} & \textbf{Optical-only} & \textbf{Large-scale dataset (3,340 images), lightweight CNN baseline} \\
			\bottomrule
		\end{tabular}
	\end{table*}
	
	\textbf{(5) Quantitative performance comparison.}

	We evaluate the proposed MCD-Net against several baselines to assess the effect of backbone architecture and attention integration.
	Specifically, we implement DeepLabV3+ networks with three different encoder backbones—MobileNetV2, ResNet152, and Xception—each trained both with and without the Convolutional Block Attention Module (CBAM)~\cite{woo2018cbam}.
	The proposed configuration, MCD-Net = MobileNetV2 + CBAM + DeepLabV3+, is designed as a lightweight yet expressive baseline for optical-only moraine segmentation.
	
	In addition, classical segmentation architectures such as U-Net~\cite{ronneberger2015unet} and PSPNet~\cite{zhao2017pyramid} are reimplemented for comparative benchmarking under identical experimental conditions. 
	All models are trained from scratch on the MCD dataset without external pretraining, ensuring a fair evaluation of generalisation capacity.
	
	Results in Table~\ref{tab:sota_comparison} show that MCD-Net surpasses conventional architectures by 2–8\% in mIoU while remaining the most parameter-efficient.
	
	\begin{table*}[t]
		\centering
		\caption{Quantitative comparison of MCD-Net with existing state-of-the-art segmentation models on the MCD dataset.}
		\label{tab:sota_comparison}
		\begin{tabular}{lccccc}
			\toprule
			\textbf{Model} & \textbf{Backbone} & \textbf{mIoU (\%)} & \textbf{Dice (\%)} & \textbf{Recall (\%)} & \textbf{Pixel Acc. (\%)} \\
			\midrule
			U-Net \cite{ronneberger2015unet} & --- & 58.0 & 67.6 & 63.5 & 91.4 \\
			PSPNet \cite{zhao2017pyramid} & ResNet50 & 60.2 & 67.5 & 66.7 & 91.5 \\
			DeepLabV3+ \cite{chen2018deeplab} & ResNet152 & 60.2 & 70.8 & 70.0 & 90.1 \\
			DeepLabV3+ \cite{chen2018deeplab} & Xception & 56.5 & 66.2 & 63.5 & 90.2 \\
			\textbf{MCD-Net (ours)} & MobileNetV2+CBAM & \textbf{62.3} & \textbf{72.8} & \textbf{70.9} & \textbf{91.2} \\
			\bottomrule
		\end{tabular}
	\end{table*}

	MCD-Net not only establishes a reproducible optical-only benchmark but also matches or exceeds the accuracy of DEM- or SAR-assisted models such as MorNet. 
	This validates the dataset’s suitability for fair benchmarking and the model’s generalisability for lightweight geomorphological applications.
	
	\section{Results and Analysis}
	\label{sec:analysis}
	
	\subsection{Overall Performance}
	Across all experiments, the MobileNetV2 backbone provides the best balance between segmentation accuracy and computational efficiency.  
	While deeper networks theoretically capture richer context, they exhibit overfitting and unstable generalisation, producing fragmented or noisy predictions.  
	MCD-Net’s moderate complexity enables consistent boundary detection while maintaining efficient inference, establishing a strong baseline for future moraine-mapping studies.
	
	\subsection{Error and Robustness Analysis}
	
	\begin{figure*}[t]
		\centering
		\includegraphics[width=0.95\linewidth]{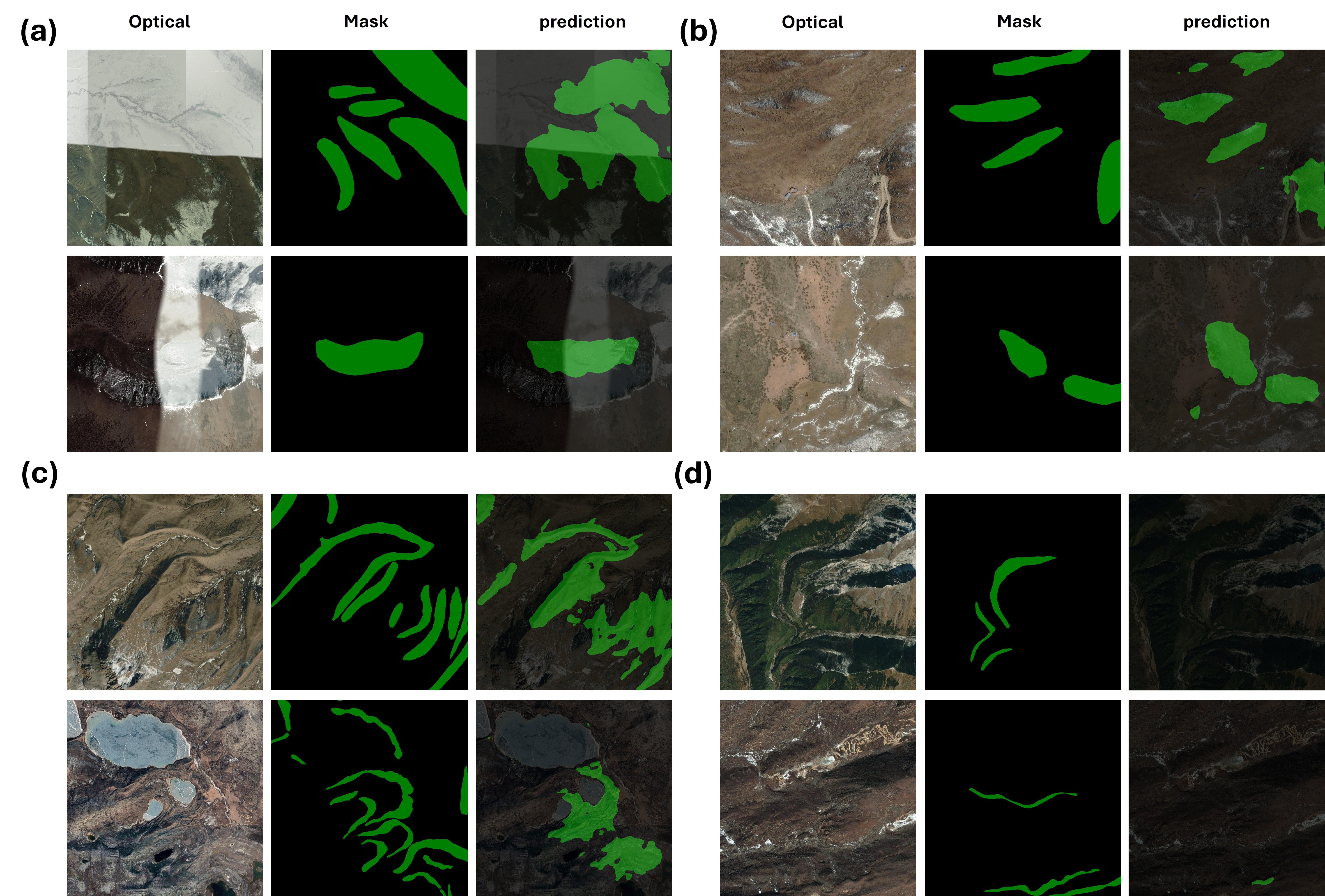}
		\caption{Comprehensive analysis of MCD-Net performance across four challenging moraine segmentation scenarios. Each scenario presents two representative examples with original imagery (left) and corresponding model predictions (right). 
			\textbf{(a) Multi-source image mosaicking artifacts}: Performance correlates strongly with regional color consistency, with severe degradation in areas affected by sensor-specific artifacts and illumination variations. 
			\textbf{(b) Morphologically degraded moraines}: Moderate success in detection but significant boundary ambiguity in heavily eroded and vegetation-covered regions. 
			\textbf{(c) Small-scale dense clustering}: Instance merging phenomenon where closely spaced moraines are incorrectly segmented as continuous features. 
			\textbf{(d) Isolated small-scale targets}: Near-complete detection failure due to insufficient pixel-level representation and feature scarcity.}
		\label{fig:challenging_scenarios}
	\end{figure*}
	
	To comprehensively evaluate the robustness and limitations of MCD-Net, we systematically analyzed its performance across four representative challenging scenarios that reflect real-world complexities in moraine mapping. Figure~\ref{fig:challenging_scenarios}
	presents both successful and failed cases, providing critical insights into the model's operational boundaries.
	
	\subsubsection{Multi-source Image Mosaicking Artifacts}
	
	The first scenario addresses challenges arising from heterogeneous data sources. As shown in Figure~\ref{fig:challenging_scenarios}(a), the model demonstrates strong performance in regions with consistent spectral characteristics but fails dramatically in areas affected by significant color discrepancies from image stitching. Despite training on datasets containing substantial mosaicked imagery, the model struggles to learn feature invariance under extreme illumination and color variations. This limitation reflects the inherent sensitivity of deep neural networks to severe domain shifts, which remains challenging to overcome even with extensive training data. The performance degradation in overexposed regions particularly highlights the vulnerability of optical approaches to extreme lighting conditions and inter-sensor spectral response differences.
	
	\subsubsection{Morphologically Degraded Moraines}
	
	Scenario Figure~\ref{fig:challenging_scenarios}(b) examines moraines affected by post-depositional processes including erosion, vegetation colonization, and sediment mixing. While MCD-Net achieves reasonable detection rates for moderately degraded features, it struggles with heavily altered moraines where spectral signatures converge with surrounding terrain. The primary challenge lies in the fundamental ambiguity between weathered moraine materials and adjacent colluvial or alluvial deposits, which share similar mineralogical compositions and surface textures. This underscores the limitation of relying exclusively on optical characteristics without complementary topographic cues.
	
	\subsubsection{Small-scale Dense Clustering}
	
	The third scenario focuses on moraine complexes characterized by high spatial density and small individual sizes. As illustrated in Figure~\ref{fig:challenging_scenarios}(c), the model tends to merge adjacent small moraines into larger segments, reflecting limitations in instance-level discrimination. This behavior arises from the inherent trade-off in CNN architectures between receptive field size and spatial resolution. While larger receptive fields capture contextual information beneficial for moraine body detection, they simultaneously reduce the model's capacity to resolve fine boundaries between closely spaced targets.
	
	\subsubsection{Isolated Small-scale Targets}
	
	Scenario (d) represents the most challenging condition, where moraines appear as isolated small features with minimal pixel representation. The near-complete failure in these cases reveals fundamental resolution constraints of current optical imagery for moraine mapping. When target features occupy only a few pixels, they lack sufficient texture and shape information for reliable discrimination. Moreover, the extreme class imbalance (moraine pixels representing <10\% of total area) further biases the model against detecting such sparse features, as false negatives incur minimal penalty during training.
	
	These analyses collectively demonstrate that while MCD-Net provides a robust baseline for moraine segmentation under favorable conditions, significant challenges remain in handling real-world complexities. The findings highlight the need for multi-modal approaches incorporating topographic data, advanced attention mechanisms for small object detection, and domain adaptation techniques to address sensor heterogeneity.

	\subsection{Discussion}
	The results demonstrate that high-resolution optical imagery, despite its limitations, can be used to reliably delineate moraine bodies across diverse geomorphological settings. The MobileNetV2-based MCD-Net outperforms deeper backbones such as ResNet152 and Xception, confirming that lightweight architectures generalise more effectively in data-limited geomorphological contexts. The modest yet consistent improvement from CBAM further indicates that selective feature refinement benefits compact models.
	
	The most challenging scenarios arise in regions affected by illumination variability, vegetation cover, mosaicking artefacts, and small-scale morphological degradation. These conditions impose inherent limits on optical-only approaches and highlight the importance of data quality and spatial resolution. Nevertheless, the proposed MCD dataset offers a substantially larger and more diverse benchmark than previous moraine-specific datasets, enabling more comprehensive training and evaluation of segmentation models.
	
	Future work may explore higher-resolution UAV imagery, domain-adaptive learning strategies to mitigate spectral variability, and the incorporation of DEM or SAR data where available. The baseline and dataset established in this study provide a reproducible foundation for these developments and open the door to more specialised architectures for small and subtle geomorphological features.

	\section{Conclusion}
	\label{sec:conclusion}
	
	This study establishes the first reproducible benchmark for optical-only moraine segmentation. 
	We curated a dataset of 3,340 annotated high-resolution images from Sichuan and Yunnan, China, and proposed MCD-Net, a lightweight DeepLabV3+ variant that integrates MobileNetV2 and CBAM. 
	The results demonstrate that optical imagery alone can provide reliable delineation of moraine bodies, while ridge extraction remains constrained by the resolution limits of current satellite data.
	
	Several overarching conclusions can be drawn. 
	First, lightweight architectures outperform deeper models in both accuracy and computational efficiency, confirming their suitability for data-scarce geomorphological applications. 
	Second, attention mechanisms offer meaningful improvements only for compact backbones, emphasising the importance of architecture-aware attention design. 
	Third, although optical imagery forms a scalable and cost-effective basis for moraine mapping, more detailed geomorphological interpretation—particularly ridge-level analysis—will require integration with higher-resolution or multi-source data.
	
	The MCD dataset and MCD-Net together provide a reproducible foundation for future research in glacial geomorphology. 
	By coupling open data with an efficient baseline model, this work establishes a transparent framework for future optical and multi-modal studies, advancing automated moraine monitoring and contributing to improved assessments of landscape change and climate impact in alpine environments.

	
	
	
	
	


\end{document}